%% file: ipsn15.tex
\documentclass[11pt]{article}

\usepackage{graphics}
\usepackage{graphicx}
\usepackage{amsmath}
\usepackage{amssymb}
\usepackage{url}
\usepackage[usenames,dvipsnames]{color}
\usepackage{epstopdf}
\usepackage{dsfont}
\usepackage{enumerate}
\usepackage{cite}
\usepackage{bbm}
\usepackage{color}
\usepackage[centering, margin={1in, 1in}, includeheadfoot]{geometry}
\usepackage{epstopdf}
\usepackage{dsfont}
\newcommand{\E}{\mathcal{E}}
\newcommand{\G}{\mathcal{G}}
\newcommand{\D}{\mathcal{D}}

\newcommand{\A}{\mathcal{A}}

\newcommand{\mS}{\mathcal{S}}

\newcommand{\M}{\mathcal{M}}

\newcommand{\beq}{\begin{equation}}
\newcommand{\eeq}{\end{equation}}

\begin{document}
\title{Dynamic Topological Mapping with Biobotic Swarms \footnote{This work was supported by the National Science Foundation under award CNS-1239243.}}
\author{
{Alireza Dirafzoon, Alper Bozkurt, and  Edgar Lobaton \footnote{ \noindent Department of Electrical Engineering, {North~Carolina~State~University}. Email: 
 \{adirafz, alper.bozkurt, ejlobato\}@ncsu.edu} }}
\date{}
\maketitle

\maketitle
\input{Abstract.tex}
\input{Introduction.tex}

\input{RelatedWork.tex}

\input{Background.tex}
\input{ProblemStatement.tex}
\input{DynamicMapping.tex}
\input{SimulationResults.tex}
\input{Conclusion.tex}

\bibliographystyle{ieeetr}
\bibliography{qual,cinema}

\end{document}

%% file: Abstract.tex
\begin{abstract}

In this paper, we present an approach for dynamic exploration and mapping  of  unknown environments using a swarm of biobotic sensing agents, with a stochastic natural motion model and a leading agent (e.g., an unmanned aerial vehicle). The proposed robust mapping technique constructs a topological map of the environment using only encounter information from the swarm. A sliding window strategy is adopted in conjunction with a topological mapping strategy based on local interactions among the swarm in a coordinate-free fashion to obtain local maps of the environment. These maps are then merged into a global topological map which can be visualized using a graphical representation that integrates geometric as well as topological feature of the environment. Localized robust topological features are extracted using tools from topological data analysis.  Simulation results have been presented to illustrate and verify the correctness of our dynamic mapping algorithm.
 
\end{abstract}

%% file: Introduction.tex
\section{Introduction} \label{sec:intro}

Sensor networks with their broad application in  
  mapping and navigation  \cite{Kotay06}, habitat monitoring  \cite{Mainwaring2002}, exploration, and search and rescue  \cite{Kantor2003},  have attracted a lot of attention in recent decades.   Mobile sensor networks 
    offer the flexibility to adapt with dynamic environments. 
%
 In some applications, however,  the amount of information that could be sensed or transferred by the agents is limited.  This motivates the design of distributed systems composed of simple agents with  minimal  sensing requirements\cite{gandhi08}.

\begin{figure}[t]
  	\centering
  	\includegraphics[width=0.5\linewidth]{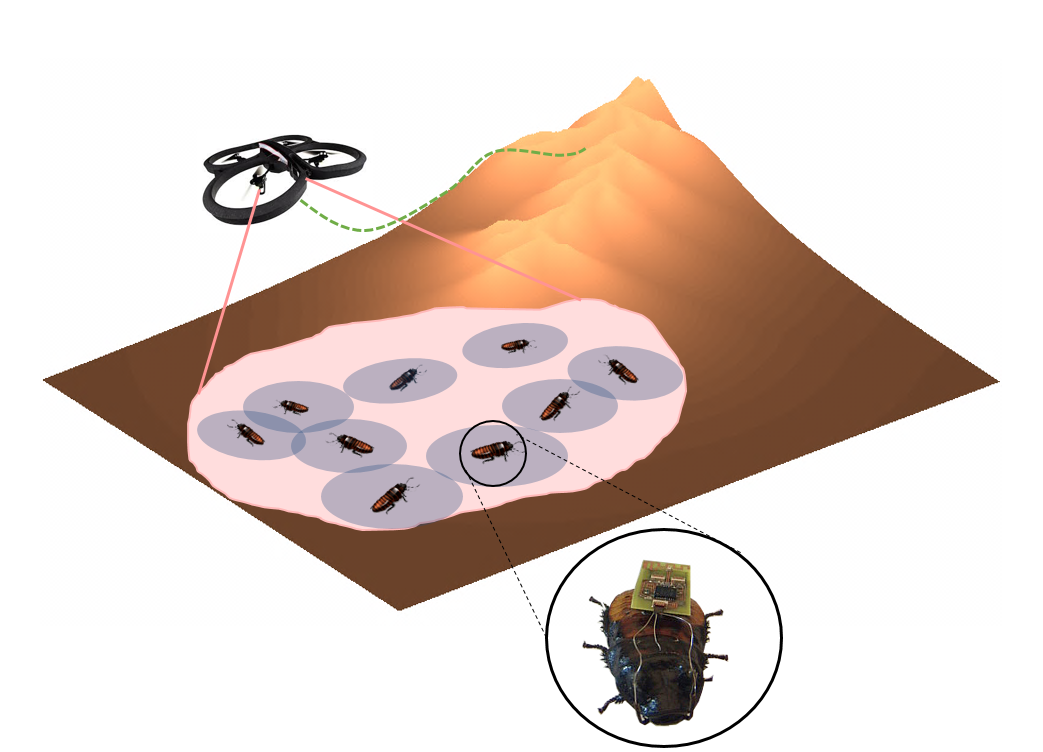}
  	\caption{ \small {A physical environment with a swarm of biobotic (cyborg insect)
  			agents\cite{Latif2012} on the ground with their corresponding local sensing neighborhoods and an aerial leading agent. 
  		}}
  		\label{fig:front}
  	\end{figure}

As a motivating example, imagine a scenario in which an earthquake has left several individuals trapped under the ruins of collapsed buildings. Sending teams of human rescuers to search and extract survivors may put the rescuers and the survivors in danger. Instead, we can choose to send multiple autonomous robotic agents to accomplish this search and rescue operation \cite{Cai}. Since large robots may not be able to explore all desired locations due to size and safety limitations, a team of small autonomous agents can be considered (e.g., biobotic insects \cite{Boz13,Latif2012} or biologically inspired milli-robots \cite{Haldane13}; see Fig. \ref{fig:front} for illustration).  The first requirement for such system would be to explore and build a map of the environment and then localize anyone in need of assistance. This is but one scenario in which a swarm of small robotic agents can be used in unstructured emergency-response situations for mapping and source-localization purposes.

The mapping task in such situations becomes extremely challenging due to hardware limitations and the unstructured nature of the environment. Power and computational resource constraints prohibit us from using continuous control schemes for the agent's locomotion and from using on-board imaging techniques for their localization. Furthermore, since some locations can be indoor or even underground in cluttered environments, signal propagation based localization (e.g., GPS, or computing signal strength or time of flight) may be unreliable, and odometry information might include a high amount of uncertainty due to irregular conditions of the terrain. Therefore, traditional mapping and exploration techniques such as SLAM \cite{SLAM} would not perform well under these adverse conditions.

   \begin{figure}[tbph]
         \centering
       \includegraphics[width=0.7\linewidth]{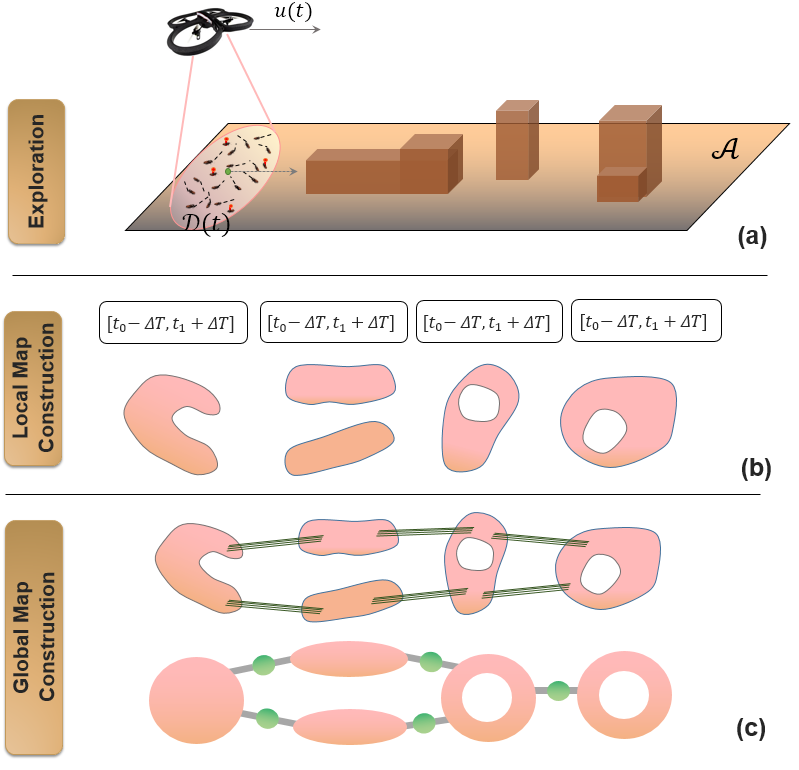}
       \caption{\small{Illustration of the physical environment to be explored with the swarm of agents performing the exploration and mapping, where $ \A $ is the area of interest and $ \D(t) $ denotes the covered domain by the aerial agent flying with the velocity of $ u(t) $ at time $ t. $} }
       \label{fig:dynam-cov}
    \end{figure}

Since obtaining an accurate metric map of the environment may not be possible in these scenarios,  we are interested to construct topological maps of unknown environments using biobotic  mobile sensor networks under the constraint of limited sensing information. 
Our approach to mapping in such scenarios is to extract a sketch of the environment instead of a fully-detailed map. 
This sketch is composed of robust topological information that is obtained from a minimal amount of sensing. Instead of providing continuous control feedback to the agents, we explore how stochastic motion models and weak encounter information can be exploited for mapping. These strategies for motion and sensing accommodate for the hardware limitations of the platforms under consideration. We employ tools from algebraic topology to extract spatial information of the environment based on neighbor to neighbor interactions among the agents with no need for localization data. This information is used to build a map of persistent topological features of the environment, which will contain only robust features.

In this paper, we extend the approach introduced in \cite{Dirafzoon2013, Dirafzoon2014},    where a notion of \textit{encounter metric} is built upon encounters among the agents, and uses topological data analysis (TDA) \cite{Edel2000} to  extract  features such as connected components and holes in the environment. This metric was improved in \cite{Dirafzoon2014} to obtain a more accurate point cloud representation of the unknown environment, by using  a density based subsampling procedure 
as well as a classification technique to provide  a robust quantitative representation of  features in the point cloud data.
Although the aforementioned approach works quite well in simple scenarios, its performance in accurately classifying features degrades for more complex environments, with growing number or  size  of persistent features in the environment\cite{Dirafzoon2014}. 
 Additionally, the complexity of the algorithm grows quadratically with respect to the number of observations required to properly estimate a map of the scene. This makes the procedure computationally too expensive for larger and more complex environments.

Therefore, in this paper, to avoid increasing the computation cost and to obtain more accurate and robust maps for larger environments with more complex features, we propose a dynamic exploration and mapping strategy by addition of a \textit{leading agent} (see Fig.~\ref{fig:front} and \ref{fig:dynam-cov}).  This approach can be described as a two level procedure: 
Globally, at  a higher level, the aerial agent performs a dynamic sweeping coverage of the area of interest by leading the center of the swarm to move towards a desired direction. 
Locally, the swarm of boibotic agents explore their assigned coverage domain.
Information gathered over smaller time windows are exploited to build local maps  of the domains covered and then merged into a global map (Fig. \ref{fig:dynam-cov}). 



%


%

The remainder of the paper is organized as follows: Section  \ref{sec:related-work} overviews the related work to this study. A concise background on the tools from topological data analysis used in this work is provided 
in section \ref{sec:Background}; section \ref{sec:ProblemStatement} describes the problem under study, including mobility and sensing characteristics of the ground biobotic agents as well as the leading agent; in section  \ref{sec:Method} we present our methodology for dynamic exploration and topological mapping with minimal sensing information; section \ref{sec:SimulationResults} validates our approach using numerical simulations; finally, conclusions and future work are discussed in section \ref{sec:conclusion}.

%% file: RelatedWork.tex
\section{Related Work}\label{sec:related-work}

The standard localization and mapping techniques, originated in the SLAM algorithm \cite{SLAM}, estimate the position of moving agents based on a set of observations from landmarks in the environment as well as sensing information from the moving agents (e.g. odometry, IMU, or vision) to estimate the agents' motion\cite{slam06}. 
As an example, a lot of recent studies on localization and mapping, has focused on visual or RGB-D SLAM \cite{visualslam, rgbdslam}. The primary assumption in these approaches is, however, that the exploring agents are equipped with advanced sensing devices such as depth sensors or camera.  In our problem, on the other hand, we consider a case where such sensing information is unavailable (e.g. biobotic or micro-robots), which will result in the failure of standard methods. 

  Methods from computational topology, on the other hand, can provide tools to    
  extract topological features from data sets without requiring coordinate information. This makes them more suitable for scenarios in which weak or no localization is provided.
Topological frameworks, which apply tools from algebraic topology, have been recently used for a variety of applications including sensor networks and robotics. Particularly,  these methods have been used  for applications such as coverage  and hole detection in sensor networks  \cite{Tahbaz10, Ghrist, Muhammad2007, Muhammad072, DeSilva2006}, motion planning  \cite{Kim13}, 
 localization \cite{Rob13, Rob12, Mike12}, and camera network coverage\cite{Lobaton2009a}.

Walker \cite{Walker2008} employed persistent homology to compute topological invariants from encounter data of the mobile nodes in Mobile Ad-Hoc networks in order to infer global  information regarding the topology of a physical environment, but the nodes are assumed to follow a simple mobility model on a graph.  
A topological version of localization via signals of opportunity has been studied in \cite{Mike12}. However, this approach is mostly focused on localization problem although it could also detect coarse features of the environment in the context of mapping. Furthermore, it requires the existence of transmitters in the environment that provide such signals of opportunity.
Topological localization and mapping has been also considered lately in \cite{Rob13} and \cite{Rob12}. Although these  are akin in the context of using  topology for representation of physical structures in the environment with mobile agents, they differ with our work in the sense that they assume the existence of an IMU or a motion model for prediction of the agents' states, as well as fixed landmarks on the boundaries of structures (e.g. buildings) to be used for observation. In this paper, however,  the mapping is intended to be performed using local \textit{encounters} among the agents and not relying on external resources (signals of opportunity or fixed landmarks), as they might get out of order as a result of  disaster scenario conditions, nor on odometry information. 

In \cite{Ghrist, Silva2007},  topological characterization of coverage and hole detection in sensor networks using only proximity information of the nodes  within a neighborhood was examined. 
%
However, these studies focused on stationary networks, and  mainly concerned about the coverage holes in the sensing domain of the  network rather than the  characterization of the physical environment itself. 


There exists also a notion of \textit{topological SLAM} in the literature (e.g. \cite{Choset01, Tully12}), which differs from our work in the sense that the context of topology has been considered as a combinatorial representation of the free space of the environment.

%% file: Background.tex
\section{Background} \label{sec:Background}

%
%

\subsection{Toppological Data Analysis}

A brief introduction of some of the basic concepts in computational topology and persistent homology 
used for data analysis in this paper 
is presented here.  A comprehensive review of the  topic can be found in \cite{Edels10}. 

Topological data analysis (TDA), introduced in \cite{Edel2000},  is  a new field of study which employs tools from persistent homology theory \cite{Edels08}
to obtain a qualitative description of the topological attributes and visualization of  data sets sampled  from  high dimensional point clouds. 


  \begin{figure}[tbph]
      \centering
   \includegraphics[width=0.4\linewidth]{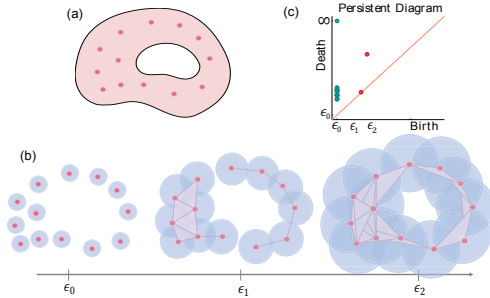}
      \caption{ \small {Topological persistence: (a) An example of a topological space $ M $ with a sampled point cloud $ X $, (b) a filtration of simplicial complexes over $ X$, (c) the corresponding persistence diagram with dgm$ _0(X) $ in blue and  dgm$ _1(X) $ in red, highlighting persistence of connected components and holes over the scale, respectively. }}
      \label{fig:filtr}
   \end{figure}
A\textit{ point cloud} can be thought of as finite samples taken from a density map which may include noise.  
TDA
represents  the prominence of features in the point cloud in terms of 
a compact representation of the multi-scale topological structure
 called \textit{persistence diagrams}\cite{Edels10}. It reduces the dimensions of data by construction of a filtration of  combinatorial objects, which can represent geometrical and topological features of the data set at specific scales, and the nodes at this complex can be considered as corresponding clusters in the data. 


One of the well-known techniques widely used in topological data analysis is persistent homology, 
which deals with the way that objects are connected. Topological structures of a  space $M$ are summarized 
as a compact representation in the form of so-called  Betti numbers, which are ranks of  topological invariants, called \textit{homology groups}. The $n$-th Betti number, $ \beta_n $ measures the number of $n$-dimensional cycles in the space (e.g. $\beta_0$ is the number of connected components and $\beta_1$ is the number of  \textit{holes} in the complex).
The space $M$ usually is not directly accessible but  a  sampled version of it, ${X,}$ can be used for computations. This sample  is represented as a\textit{ point cloud},  a finite set of  points equipped with a metric, which can be defined by pairwise distances between the points (distance function). The distance function can be defined based on the coordinates of the points in their embedding manifold, or can be constructed by specifying pairwise distances between points independently from their (possibly unknown) coordinates. 

A standard method to analyze the topological structure of a point cloud  is to map it  into  combinatorial objects called {\it  simplicial complexes}. 
One way to build these complexes is to select a scale $\epsilon$, place balls of radius $\epsilon$ on each vertex, and construct simplices based on their pairwise distance relative to $\epsilon$. A computationally efficient complex, called the \textit{Vietoris-Rips complex} \cite{Erin10}, consists of simplices for which the distances between each pair of its vertices are at most $\epsilon$. 
A sequence of complexes, called a filtration ${X}(\epsilon)$,  then can be obtained by increasing $\epsilon$ over a range of interest,  with the property that 
if  $ t< s $ then  ${X}(t)\subset {X}(s)$. 
Persistent homology, computes the  values  $\epsilon$ for which the classes of topological features  appear ($b^i_n$) and disappear ($d^i_n$) during filtration,   referred to as  the birth and death values of the $i$-th class of features in dimension $ n $. 
 This information is encoded into persistence intervals $[b^i_n, d^i_n] $ or as a multiset of points $ (b^i_n, d^i_n) $, called a persistence  diagram, dgm$_n $($ {X} $). Algorithms for computation of persistent homology  can be found in \cite{Edel2000, Zl04}. 
An example of a topological space $M $ together with sampled data ${X} $ and corresponding filtration over $ \epsilon $ is shown in Fig. \ref{fig:filtr}. The persistent diagram infers the existence of one connected component and one persistent hole. 

%% file: ProblemStatement.tex
\section{Problem Statement} \label{sec:ProblemStatement}

Consider a collection of  mobile sensing agents moving stochastically on a bounded domain of interest $\mathcal{A} \subset \mathbb{R}^2$.   Each ground agent is distinguished by its unique  ID, and its  motion dynamics in the bounded space mimic the movements of a biobotic insect. In addition to the swarm of ground agents, there exists a leading agent (e.g. a quad-copter or a biobotic moth), which plays the role of a swarm leader in order to herd the ground agents to explore all parts of the desired environment over time. It serves as an access point to establish a local network for communication with the swarm in order to send control commands and receive required information (Fig. \ref{fig:dynam-cov}).  

The ground agents are provided with weak localization information, i.e. neighboring agent identification within a specific range are the only information used, and no coordinate information or other information is provided to them. For example, odometry and inertial measurements are considered to be too unreliable due to the uneven and unstructured terrains present in our application scenario.  However, they can communicate wirelessly to the leading agent within its limited communication range. Alternatively, due to weak transmitters on the ground agents, we could just consider a one-way communication from the aerial agent to the swarm.

\begin{figure}[tbph]
         \centering
       \includegraphics[width=0.5\linewidth]{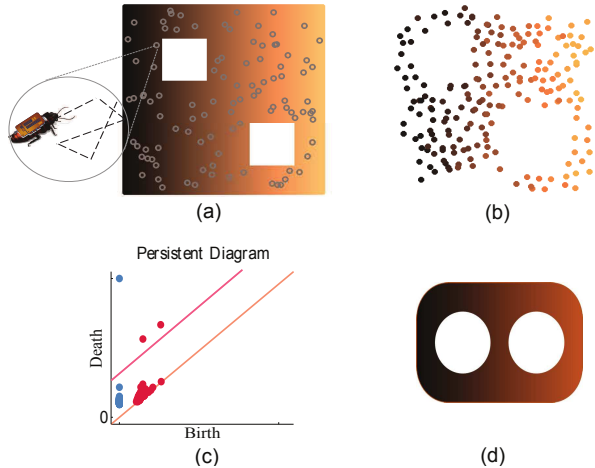}
       \caption{\small{Topological Mapping: (a) physical environment, (b) estimated point cloud from encounter information,  (c) persistence diagram highlighting connected components in blue and holes in red dots (significant features can be distinguished from noise using an appropriate threshold depicted as a red line), and (d) sketch of the environment. }}
       \label{fig:top-map}
    \end{figure}

Given this minimal sensing scenario, we aim to build a sketch of the unknown scene, which includes information about topological and geometrical features of the environment $\A$. We refer to such sketch as the \textit{topological map}. Fig. \ref{fig:top-map} illustrates the idea of such topological mapping for a square shaped physical environment including two obstacles. Our description incorporates a point cloud representation obtained from local interactions between agents (Fig. \ref{fig:top-map}(b)), a list of robust features captured by topological persistence diagrams (Fig. \ref{fig:top-map}(c)), and a physical sketch of the environment (Fig. \ref{fig:top-map}(d)).

The previous example demonstrates the approach proposed by the authors in \cite{Dirafzoon2014}, in which stochastic agents completely explore a bounded domain in order to construct a topological map. These results will be the building blocks for the approach proposed in this paper, in which an external agent is used to lead the swarm in order to allow for mapping and exploration of large and more complex environments. We assume the existence of a leading agent (e.g., an aerial vehicle) which sweeps the entire domain. The stochastic agents are assumed to stay within a region $\D(t) \subset \A$ centered around the leading agent. As the swarm of agents travel through the environment, a dynamic map of the scene is built by having the agents share information about their interactions. In order to provide enough encounters between agents to be gathered, the leading agent must travel at an appropriate speed. Furthermore, we require that the dynamic coverage of the stochastic swarm to cover the domain of interest. That is,
\begin{equation}
 \A \subset \cup_{t\in [0,T]} \D(t), 
\end{equation}
where $[0,T]$ is the observation time interval.

\subsection{Motion Model of the Biobotic Swarm }\label{subsec:model}


The mobility  model of the ground mobile sensors is adopted from the probabilistic movement model of cockroaches, described in \cite{Jeanson2003} ,which can be mainly described by their individual and group behaviors. 

The  individual behavior of  cockroaches can be described mainly by two behaviors, namely diffusive \textit{random walk} (RW) and \textit{wall following} (WF).  
 In this paper, for the sake of simplicity, we skip the group behavior and wall following and only consider agents who stochastically move according to the  RW model.  This is consistent with the behavior of cockroaches \cite{Jeanson2003}, which can be exploited as a \textit{natural} mode of operation for biobotic systems \cite{Latif2012}.
 
 The {random walk} is modeled as piecewise linear movements with fixed orientation, interrupted by isotropic changes in direction, and constant average velocity of $v_m$. The length of the line segments  are selected randomly from an exponential distribution with characteristic length $\lambda$. Changes in direction are also triggered by collision detection with other agents or obstacles in the scene.

  Moreover, it is known that during their RW motion, the agents  probabilistically \textit{stop} for some period of time and then continue their movement\cite{Jeanson2003}. 
 The biobotic  insects 
 are equipped with a \textit{controlled motion} mode  aside with their \textit{natural motion} mode. In the controlled motion mode, the insects are capable of being controlled wirelessly to perform \textit{left} and  \textit{right} turns or to be commanded to \textit{start} or \textit{stop} their motion. Consistent with this controlled mode, we assume that,  
  aside from \textit{random walk}, the agents can be in a \textit{static} mode of operation in which they do not move but sensing and communication are still possible. The only control that we assume on their motion is switching between \textit{static} and \textit{random walk} modes.

Note that the average velocity of the swarm is determined by the average velocity of the flying leader, $ \bar{u} $. 
In order for the biobots to fully explore and sample the current coverage domain of the leader with high enough density, the average velocity of the random walk motion of the swarm agents needs to be high enough compared to the swarm velocity.

\subsection{Sensing and Communication }

The sensing and communication model for the agents is inspired by limited sensing capabilities of the biobotic insects, which incorporate a wireless transmitter and receiver provided by a system-on-chip based ZigBee enabled wireless backpack system inserted into their bodies\cite{Latif2012}. Each agent is distinguished by its unique ID, and assumed to have a limited sensing capability, with a detection radius $r_d$, which
defines the region in which it can identify and communicate with other agents. It is assumed that the swarm can only detect signal strength from the leading agent, which can be used to maintain the agents within a bounded domain $D(t)$ centered around the leading agent at time $t$.

 Moreover,
each agent can detect the presence of an obstacle and avoid
collision by changing direction within its collision detection
region.   The agents are able to
record their \textit{encounters} with each other in accordance with
the corresponding times as encounter events. These encounters are in fact the main piece of information in our approach  in the construction of a metric on the samples of the physical space for the purpose of mapping. 
An encounter event takes place between two agents if their corresponding center positions fall within a distance of $r_d$. 
An encounter $ \E_i $ can be recorded by an agent as a tuple:
\begin{equation}
  \E_i= [t_{0i}, \, t_{1i}, \, \text{ID}_i^1, \, \text{ID}_i^2  ],
\end{equation}
where $[t_{0i},t_{1i}]$ denotes the time interval of the encounter, and $ \text{ID}_i^1 $ and $ \text{ID}_i^2 $ represent the IDs of the two encountering agents. We define  $T(\E_i) = [t_{0i},t_{1i}]$ as the $ i$-th event time interval, and  $ID(\E_i) = \{ \text{ID}_i^1 , \text{ID}_i^2 \}$ as its ID set .
At each time step,  we consider a subset of agents in the corresponding coverage domain $ \D(t) $ to be in the static mode, and denote by $ \mS(t) $ the set of static nodes at time $ t $. 
Furthermore,  the nodes are  able to report their status as being in a random walk or static  state. 

%% file: DynamicMapping.tex
\section{Dynamic Topological Mapping}
\label{sec:Method}

\begin{figure*}[!tb]
\centering
	\includegraphics[width=\linewidth]{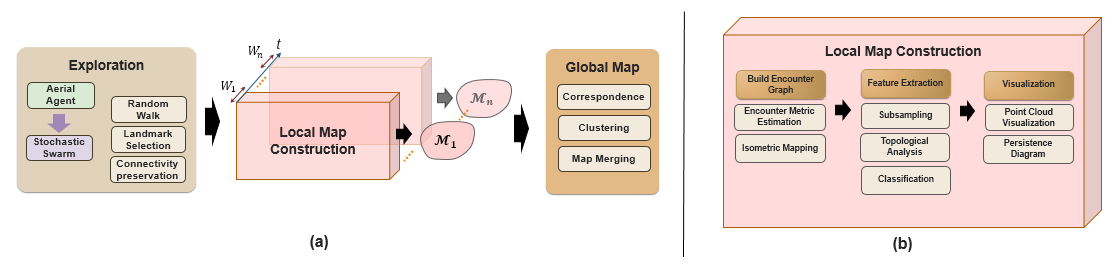}
	\caption{\small{Dynamic topological mapping (a) A block diagram illustrating several components of mapping, grouped into three major tasks: exploration, local mapping, and global merging, and (b) components of local map construction over each time window. }}
	\label{fig:over-dyn}
\end{figure*}

As described in the introduction, our objective is to construct a map of the environment in scenarios with no reliable odometry, signal strength or visual features (due to an unstructured and harsh terrain), and with limiting power constraints. For this reason, we rely only on a minimal sensing strategy that makes use of encounter information between stochastic agents (i.e., keeping track of interactions between agents that are in close proximity of each other).

In this section, we represent our methodology to extract topological and geometrical features from unknown environments. An overview of the approach is shown in Fig. \ref{fig:over-dyn}. The basic methodology makes use of the topological mapping approached introduced by the authors in  \cite{Dirafzoon2013} as a building block. As the leading agent sweeps the environment, observations over a set of windows of time are used to construct \textit{local topological maps}. Information from observations in the intersection of these windows are used to determine connections between these maps in order to stitch them together and build a \textit{global topological map}.

We define $\Delta t$ as the total amount of overlap between two time windows. This overlap needs to be sufficient in order to guarantee that enough static points and encounter events are found there to be able to properly stitch the topological maps (see section \ref{sec:GlobalMapping} for more details). The time windows from which observations will be used for the local mapping are defined as
\beq
  W_i = [t_{i-1} -\Delta t / 2, \,  t_{i} +\Delta t / 2],  i=1, \ldots, N,
\eeq
where $\{ t_i \}$ are evenly spaced, $N$ is the total number of windows, and $\bigcup_i W_i = [0,T]$ (see figure \ref{fig:dynam-cov} for an illustration).

As mentioned before, $D(t)$ is the domain of interest around the leading agent located at $x(t)$ in which the ground agents will be exploring the domain (e.g., $D(t)$ can be a disk of radius $r$ centered at $x(t)$). Hence, we can define the $i$-th domain for local mapping as
\beq
  \D_i = \bigcup_{ t \in W_i } D(t),
\eeq
which is the region in the space to be mapped locally. In order to guaranteed full coverage of the environment $\A$, we must have $\A \subset \bigcup_i \D_i$.
%

\subsection{Local Mapping}
\label{sec:LocalMapping}

This task is accomplished by following the methodology introduced in \cite{Dirafzoon2014}. The ground agents perform the exploration and mapping by simultaneous random dispersion and leader following. In order to create fixed landmarks in the environment, a subset of agents are selected as static. The process is executed by: (1) recording encounter events between agents, (2) constructing a metric based on these events, and (3) extracting robust topological features.
 
\subsubsection{Exploration and Event Collection}

The exploration  initiates with  a  \textit{dispersion} stage, where  all nodes start their motion from an arbitrary initial configuration in a moving status until dispersed throughout the environment, while maintaining their connectivity with the aerial access point. 
This is followed by a \textit{landmark selection} stage, in which a small percentage of the nodes are selected as static $ landmarks $ by modifying a standard \textit{MaxMin} landmark selection approach \cite{Silva2004}. Given the adjacency graph due to the communication range, we use its hop distance to select landmarks iteratively by maximizing its distance from the previous set of landmarks. The first agent is selected randomly. If the graph is not connected we assume that  landmark agents are selected randomly from each cluster with probability equal to the percentage of agents in each cluster. We have shown in \cite{DirafzSocg}  that incorporating a small percentage of  static nodes in the  network, as landmarks will improve our point cloud estimation of the environment to a great extent.
Once the static landmarks are selected, the rest of the agents continue their random walk motion and record their encounters with other agents.

In order to incorporate the necessary aspects of leader following, we require the ground agents to remain within $\D(t)$ for all times. We assume that the agents are capable of detecting when they leave this domain of interest, and can change their random walk strategy in order to estimate a local signal strength gradient and move in that direction. Static agents that leave $\D(t)$ would immediately switch to random walk behavior, and new landmark agents would be selected. Our method is robust in the sense that it tolerates having some agents getting separated from the swarm.

\subsubsection{Geometric Reconstruction}

As our approach is a coordinate free procedure, we need to first build an estimated metric to represent the map of the local region as a point cloud. Our metric, which we refer to as the \textit{encounter metric}, is constructed based on a set of \textit{encounter events} among the agents.

To build a distance metric on the set of encounter events,  we construct  an undirected weighted graph $\G$ with vertices corresponding to the events $\E_i$, denoted as the \textit{encounter graph}. For any two vertices $\E_i$ and $\E_j$, we include an edge $e_{ij}$ with weight
\beq
  w_{ij} = \inf\left\{ |t_1 -t_2| \; : \; t_1 \in T(\E_i) \, , \, t_2 \in T(\E_j))\right\} 
\eeq
if $ID(\E_i) \cap ID(\E_j) \neq \emptyset$. 
 We can furthermore improve our graphical model by incorporating the fact that landmark nodes are static. 
  Let $\mathcal{S}$ be the set of indices of all agents who has been static for some time period. For each $ s\in \mS $ let 
  \begin{equation}
  \mathcal{T}(s) = \{T^1(s),  T^2(s),  ...  \} 
  \end{equation}
  denote the set of all time intervals $ T^i(s) = [t^{i,b}_s,\, t^{i,d}_s ] $ for which agent $ s $ has been in a static mode.  
   Then, two events $\E_i$ and $\E_j$ with 
  \begin{equation}
 ID(\E_i) \cap ID(\E_j) = s \in \mathcal{S} 
  \end{equation}
  had to occur  geometrically at nearby locations if
  \begin{equation}
\exists k, \text{s.t.\, }   T(\E_i) \cap T(\E_j) \cap T^k(s)  \neq \emptyset.
  \end{equation}
Therefore, the corresponding weight $w_{ij}$ is set to $0$ in this case.

Note that the elements of the graph $\G$, represent estimations of pairwise distances between encounter events, i.e. $ \G(i,j)  = \hat{d}_\E (\E_i, \E_j) $. To proceed with the geometric point cloud reconstruction, we follow the\textit{ Isometric mapping (Isomap)} technique \cite{Tenenbaum2000} in \textit{manifold learning}\cite{mani-vis}.  The main idea is to estimate Geodesic distances on a manifold by the shortest path distance on the neighborhood graph of the samples taken from the manifold. 

Therefore, we build our encounter metric on $\G$,  denoted by $\D_{\G}$, 
as $[ \D_\G]_{i,j} =    d_\G(\E_i,\E_j) $
where $d_\G(\E_i,\E_j)$ represents the length of the shortest path between nodes $  \E_i,\E_j$ in $ \G $, which can be obtained from Dijkstra's or Floyd's shortest path algorithms \cite{csrl01}.
Finally, we apply  classical \textit{Multi-dimensional Scaling} (MDS)\cite{mds} to the matrix $D_\G$ in order  to find an embedding of the point cloud data in 3-dimensional Euclidean space.

\subsubsection{Topological Feature Extraction}

The set of encounter nodes together with the corresponding estimated metric defines a point cloud, the  \textit{encounter point cloud},  which is processed to construct a filtration of simplicial  complexes, denoted as \textit{encounter complexes}. This filtration is obtained by computing the Vietoris-Rips complexes \cite{Erin10} from the encounters as a function of the distance parameter $\epsilon$ (as defined in section \ref{sec:Background}). The construction of the encounter complex is done in a computationally efficient and robust way by incorporating strategies as described in \cite{Dirafzoon2014, DirafzSocg}.

We use the tools introduced in section \ref{sec:Background} in order to extract persistence diagrams from a filtration of the subsampled set of points. The filtration is based on computing the Rips complexes \cite{Erin10} and we use the Dionysus C++ library \cite{Dion2} for computation of persistent homology. The diagrams capture the persistence and  robustness of topological features in the environment. We restrict our computations to the dimensions $ 0 $ and $ 1 $ of persistence as the higher dimension are not applicable in our experiments. Robust classification of topological features in the environment is performed by using statistical learning techniques as described in \cite{Dirafzoon2014}.

\subsection{Global Mapping}
\label{sec:GlobalMapping}

So far we have constructed a set individual local maps $ \{\M_i\} $ for the covered domains $\{ \D_i \}$. The objective of this section is to provide a methodology for gluing these maps together to obtain a global one, $ \M $ for the area of interest $ \A $. Figure \ref{fig:over-dyn} illustrates our proposed approach. The idea is to use common static nodes between different map components to define a connectivity measure  and connect them based on the defined measure. To preserve topological features that cannot be described on a single local map (e.g. a very long building that will not fit within one window), we construct point clouds on the intersection of the consecutive map components ($ \M_i $ and $ \M_{i+1} $) and perform a clustering analysis to robustly identify the number of connections between the different map components.

We define the subset of static nodes joining $\M_i$ to $\M_{i+1}$ as: 
\begin{equation}
 \mS_{i,i+1}  =   \{ s  \in \mS\, : \; \exists k\;  \text{s.t. }\;  t_s^{k,b} \leq  t_i - \Delta t\,\, \text{and} \,\,\,t_s^{k,d} \geq t_i + \Delta t  \}.
\end{equation}
This is the set of static nodes whose appearance time precede the starting time of $W_{i+1}$ and their disappearance time is later than the end time of $W_i$. Furthermore, we define the set of encounter associated with these static nodes as
\begin{equation}
 \E_{\mS_{i,i+1}}  =  \{ \E_i\; : ID(\E_i) \cap \mS_{i,i+1} \neq \emptyset\}. 
\end{equation}
An encounter metric is constructed on the set $ \E_{\mS_{i,i+1}} $ to approximate a metric on the set $ \mS_{i,i+1} $. We consider a point cloud,  consisting of the set $ \E_{\mS_{i,i+1}} $ together with the k-nearest neighbors of each of its elements as a connecting point cloud between the maps $ \M_i $ and $ \M_{i+1} $. We refer to this point cloud as the \textit{inter-domain point cloud} and the equivalent metric as the \textit{correspondence metric} between two local maps $ \M_i $ and $ \M_{i+1}$. 

Note that $\mS_{i,i+1}$ correspond to connections between the two maps, but some of these connections may correspond to the same connected component within the domains $D_i$ and $D_{i+1}$. In order to identify which connections are equivalent, we perform a clustering analysis by using the encounter metric constructed on the set $ \E_{\mS_{i,i+1}} $.

As we do not know the number of clusters a priory, we employ the single linkage clustering as one of  the most famous agglomerative hierarchical clustering techniques \cite{pattern}. Hierarchical techniques provide a similarity dendrogram which can be used to classify data into clusters by applying a cut-off threshold $c$ on the similarity values. This cut-off is selected experimentally.

%% file: SimulationResults.tex
\section{Simulation Results} \label{sec:SimulationResults}
\begin{figure*}[tbph]
      \centering
  \includegraphics[width=0.9\linewidth]{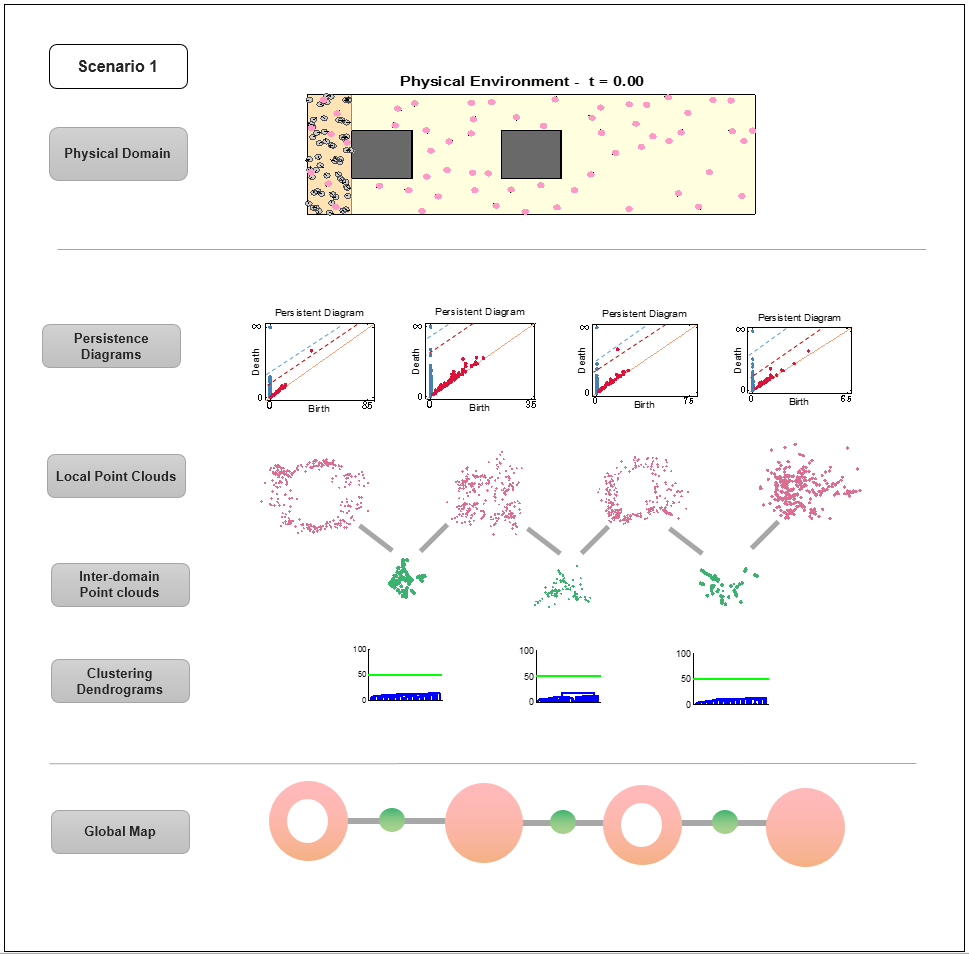}
      \caption{ \small {Simulation results for  environment scenario 1}}
      
      \label{fig:sim1}
   \end{figure*}

\begin{figure*}[tbph]
      \centering
  \includegraphics[width=\linewidth]{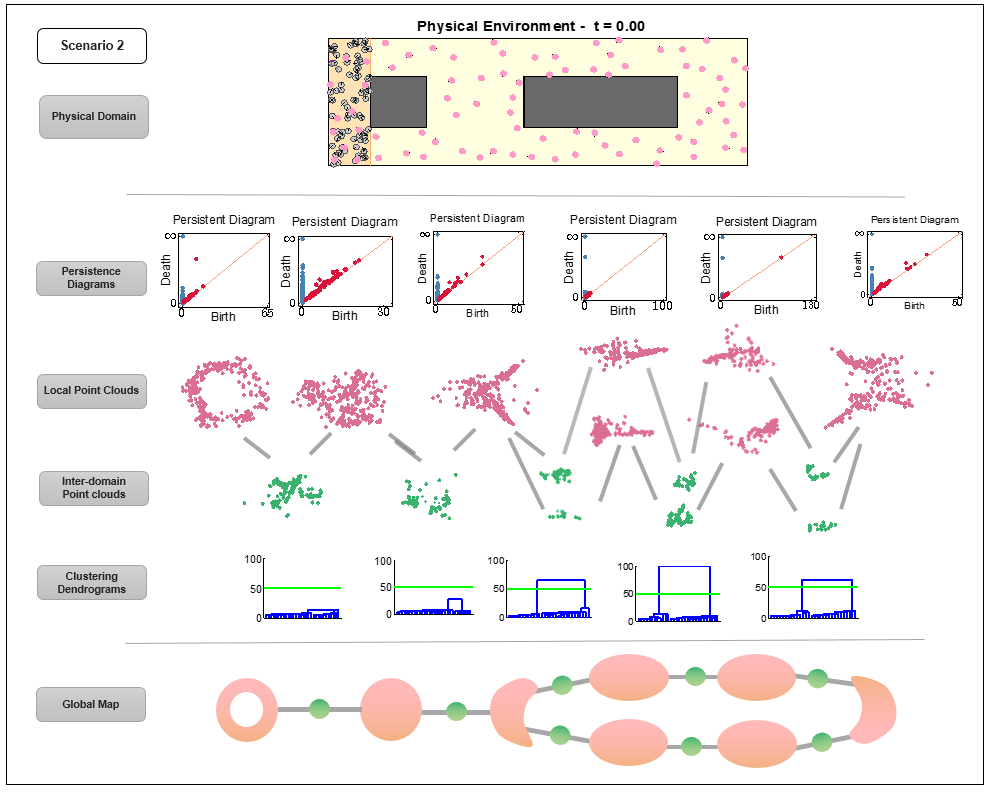}
      \caption{ \small {Simulation results for  environment scenario 2}}
      
      \label{fig:sim2}
   \end{figure*}

\begin{figure*}[tbph]
      \centering
  \includegraphics[width=\linewidth]{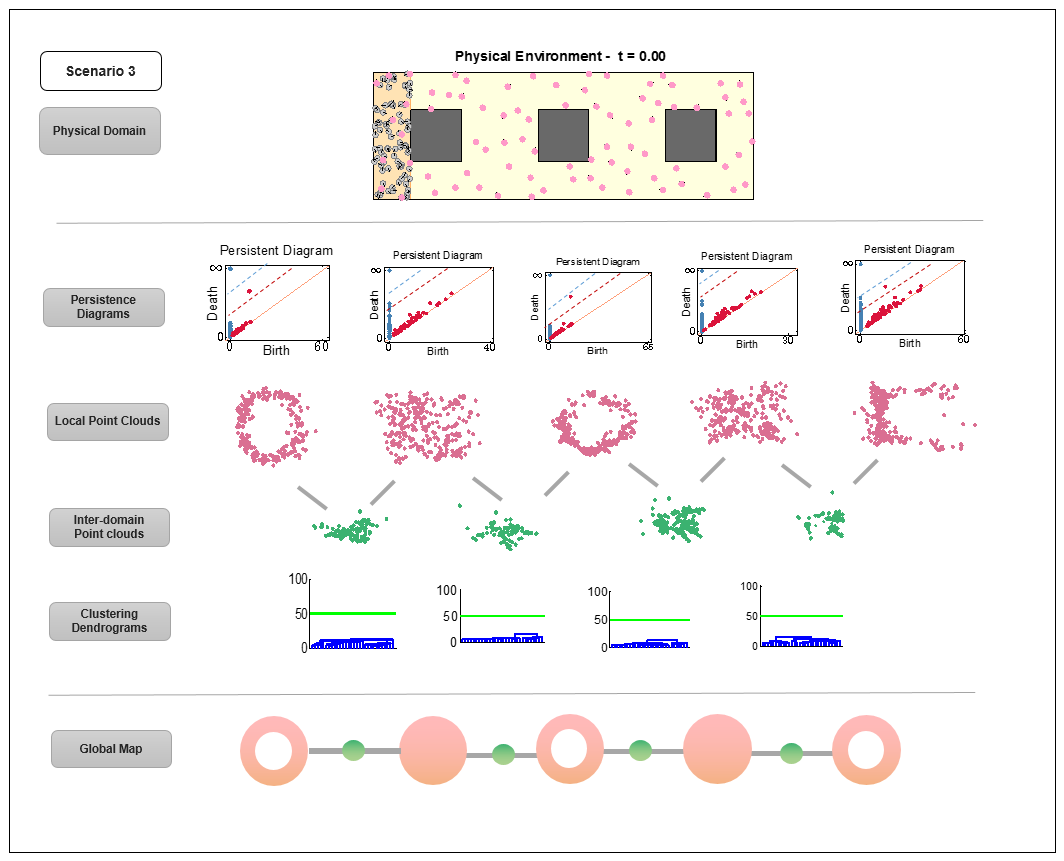}
      \caption{ \small {Simulation results for  environment scenario 3}}
      
      \label{fig:sim3}
   \end{figure*}

In this section, we present the simulation results for the verification of our dynamic topological mapping approach. We consider basic geometries in our analysis in order to provide the building blocks for a structured mapping of more complex environments.
In our simulations, we consider three different scenarios which differ in the physical layout of the area of the interest, as shown in the first rows of the Figures \ref{fig:sim1} - \ref{fig:sim3}. 
The idea behind these scenarios is to verify the successful implementation  of the proposed approach when there are large features (e.g. long structured obstacles) that cannot fit within a local coverage domain, like the one in Fig.~\ref{fig:sim2}, or scenarios like in Fig. \ref{fig:sim3}, in which although the individual features are not too big, but need to be recognized and merged properly. 

The aerial leader is considered to move with a constant linear velocity of $ v_x = 0.005 \,m/s $ in the  $ x-$direction, while the average velocity of the ground agents in their random walk model is considered to be $ v_m = 0.1 m/s $. The coverage window of the aerial vehicle, for simplicity, is considered to be a rectangular arena with the length of $ l = 3 m$, and the same width as the environment, where the dimensions of the environments are set as $ 10m \times 30m$.  
In each figure, the first row  represents the physical environment together with the initial condition of the agents (gray circles) at $ t=0 $ as well as the location of all static nodes (landmarks) over all time periods (shown as pink dots).  For each scenario, a total number of 100 agents are considered in the network, among which 10$\% $ are selected as landmarks after some initial dispersion based on the maxmin landmark selection algorithm. The detection radius of each agent for encounter and boundary recognition is adopted to be $ r_d = 1cm$. 

We divide the whole run time of scenarios 1, 2, and 3 into 4, 6, and 5 windows, respectively, and implement our proposed local map construction on gathered encounter information over each window. Note that such selection have been made for sake of better visualization of the features,  but in general we could select the same number of windows for all scenarios.  The point clouds constructed  from the estimation of local encounter metrics for different windows are presented in the third row of each figure (in pink color), in accordance with their persistence diagrams in the second rows. On each persistence diagram, 0-dimensional and 1-dimensional topological features (connected components and cycles)  are plotted as blue and red dots, respectively, and the corresponding thresholds for the persistence classifiers are sketched as dashed lines.   It can be observed the point clouds representing local maps represent the correct topological features of the corresponding local domains covered over each window. This can also be verified by the topological features classified as robust ones in the corresponding persistence diagrams. Moreover, the estimated point clouds also contain some geometric information, which could be exploited to improve the mapping accuracy in future work. 

In the next row of each the figure, the inter-domain point clouds obtained from the correspondence metric between local maps are plotted followed by the dendrograms  representing the results of the hierarchical clustering on each of them. To specify the number of clusters for each point cloud, a clustering threshold  value of 50 has been selected for all scenarios. In all cases, except the last three point clouds in scenario 2, they are classified as one cluster. On the other hand, in scenario 2, as a result of the large obstacle, local maps $ \M_4 $ and $ \M_5 $ and their corresponding inter-domain point clouds are all divided into two clusters. Finally, from the collection of local maps, and obtained correspondence metrics, we stitch the consecutive pieces of the maps together in order to find a global topological representation of the environments. The final result of such maps for each environment are sketched in the last rows of the Figures \ref{fig:sim1} - \ref{fig:sim3}.


%% file: Conclusion.tex
\section{Conclusion and Future Work} \label{sec:conclusion}

In this work, we presented a dynamic mapping approach for biobotic sensor networks under weak localization information, to obtain estimated local and global topological maps of unknown environments based on only proximity information from agents who stochastically investigate an assigned sub-domain of the environment, with few percentage of them acting as static landmarks, while tracking a leading agent by maintaining their connectivity to it in order to  cover, explore, and map an area of interest.

The topological mapping approach used to obtain local point clouds of sub-domains can be viewed as reconstruction of a manifold sampled from the space of $ \D_i \times [t_{i-1} - \Delta t, t_i +\Delta t ]$. 
Hence the accuracy of the procedure is dependent on the number of samples taken from that space, which depends on the number of agents, their average velocity, and the time assigned to explore a given sub-domain. We will explore such dependencies to come up with bounds on the values of aforementioned parameters exploiting the existing literature on topological inference from point cloud data\cite{Niy08,Niyogi11}.
Moreover, the existence and appropriate selection of static landmarks plays an important role in the construction of a distance metric under weak localization conditions. We are investigating the dependencies of the location and number of such landmarks on the estimation errors of pairwise distances, and aim to come up with criteria to choose landmarks in a more optimal and efficient manner.  
Furthermore, we are currently working on  the integration of our topological mapping approach with network SLAM algorithms with range measurements \cite{network-slam} to design a hierarchical mapping approach for disaster scenarios which require both ground and underground surveillance. 
Our future work will also include experimental results of the proposed approach with a swarm robotic platform, as well as the biobotic insects testbed in order to realize exploration and mapping with cyborg insect networks.